\documentclass[conference]{IEEEtran}
\IEEEoverridecommandlockouts

\usepackage{booktabs} 

\usepackage{tcolorbox}
\usepackage{tabularx}
\usepackage{array}
\usepackage{colortbl}
\usepackage{comment}
\usepackage{cite}
\usepackage{multirow}

\usepackage{booktabs}
\usepackage{tabularx}
\usepackage{array}

\newcolumntype{Y}{>{\raggedright\arraybackslash}X}

\usepackage{amsmath,amssymb,amsfonts}

\tcbuselibrary{skins}

\usepackage{placeins}
\def\BibTeX{{\rm B\kern-.05em{\sc i\kern-.025em b}\kern-.08em
    T\kern-.1667em\lower.7ex\hbox{E}\kern-.125emX}}

\usepackage{xcolor}
\usepackage{graphicx}
\usepackage[hidelinks]{hyperref}

\newcommand{\orcidicon}[1]{%
  \href{https://orcid.org/#1}{%
    \raisebox{-0.15ex}{\includegraphics[height=1.6ex]{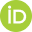}}%
  }%
}

\usepackage[subrefformat=parens]{subcaption}
\begin{document}


\author{
    Aidana Baimbetova\protect\orcidicon{0009-0001-8448-1889}$^{1}$, 
    Haruki Yonekura\protect\orcidicon{0000-0001-8184-3883}$^{1,2}$, 
    Hamada Rizk\protect\orcidicon{0000-0002-8278-8801}$^{1,2,3}$, 
    Hirozumi Yamaguchi\protect\orcidicon{0000-0003-2273-4876}$^{1,2}$\\
    $^{1}$The University of Osaka, Japan\\
    $^{2}$RIKEN Center for Computational Science, Japan\\
    $^{3}$Tanta University, Egypt\\
    \texttt{\{a-baimbetova, h-yonekura, hamada\_rizk, h-yamagu\}@ist.osaka-u.ac.jp}
}

\title{Pedestrian-Aware LLM-Driven Behavioral Planning for Autonomous Vehicles\thanks{This paper has been accepted for presentation at the 29th IEEE International Conference on Intelligent Transportation Systems (ITSC).}}

\maketitle

\begin{abstract}

Autonomous Vehicles (AVs) must make reliable decisions in dense urban environments where pedestrian behavior is variable, sometimes abnormal, and frequently unseen during training. Reinforcement learning (RL)–based AVs control systems perform well in structured traffic settings but struggle to generalize when pedestrians behave unpredictably or in scenarios outside their training distribution. Their reliance on handcrafted rewards and opaque decision further limits their suitability for safety-critical, pedestrian-rich environments.
To overcome these limitations, we introduce a Large Language Model (LLM)–based decision-making framework designed to enhance pedestrian safety and robustness under abnormal and unseen conditions. The system converts structured scene observations into natural-language reasoning prompts that allow the LLM to infer pedestrian intent, anticipate risk, and generate cautious tactical driving decisions. These decisions are then executed through a motion planner that ensures smooth and kinematically feasible control. We evaluate the framework in SUMO across multiple pedestrian-interaction scenarios, including unexpected jaywalking events and complex unseen pedestrian behaviors such as turn-back crossings, where pedestrians abruptly reverse direction during traversal. In zero-shot evaluation, the LLM-based agent achieves a 68\% collision-free success rate, substantially outperforming deep RL baselines (17.7\%). When augmented with few-shot episodic memory in a single-pedestrian scenario, performance increases to 96.0\%, outperforming a custom DQN controller (82.0\% success rate). Furthermore, in the complex unseen turn-back scenario, the few-shot memory-enabled agent demonstrates strong generalization capability, successfully adapting prior experience to previously unobserved pedestrian dynamics. The system consistently initiates earlier responses, maintains wider safety buffers, and produces interpretable, human-aligned decisions.

\end{abstract}






\begin{IEEEkeywords}
Autonomous Vehicle Control, Large Language Model, Pedestrian-aware Safety System
\color{black}
\end{IEEEkeywords}

\section{Introduction}

Autonomous Vehicles (AVs) are gaining popularity today and are anticipated to become a core of intelligent transportation in the future.  
To fully realize their potential, AVs must operate robustly in complex urban environments where decision-making depends on understanding diverse and dynamic contextual factors. Urban driving poses unique challenges due to constantly changing traffic conditions and the unpredictable behavior of vulnerable road users such as pedestrians, cyclists, and scooter riders. Effective decision-making in these environments requires adaptive reasoning about motion, intent, and social norms. 
However, many existing approaches struggle to generalize across diverse scenarios and to dynamically prioritize safety-critical elements.

Reinforcement learning (RL) methods have achieved strong results in structured driving tasks such as highway merging, intersection negotiation, and lane changing \cite{liu2022autonomous, zhou2022multi, isele2018navigating}. Yet these systems rely on extensive simulation data, long training cycles, and careful reward engineering to achieve stable performance. Their generalization remains limited: when facing novel or rare situations, RL policies often fail to adapt efficiently, making them unsuitable for unstructured real-world traffic. These limitations become particularly problematic in pedestrian-rich environments, where vehicles must respond to highly urgent and unpredictable behaviors, such as sudden trajectory deviations or unexpected jaywalking. Moreover, the low interpretability of RL policies and their dependence on task-specific tuning hinder transparent evaluation and human oversight.

Recent advances in Large Language Models (LLMs) offer a different capability: contextual reasoning beyond the training distribution. LLM-based systems have demonstrated the ability to integrate multi-modal information, interpret high-level goals, generalize from few examples, and produce interpretable justifications \cite{wang2023drivemlm, wang2023accidentgpt, huang2024making, mao2023agent, lu2025convoyllm}. 
Recent efforts have also explored LLM-assisted autonomous mobility beyond conventional driving, including cross-city autonomous driving with meta-learning and LLMs~\cite{10.1145/3748636.3766536} and multimodal LLM-driven robot navigation in dynamic human environments ~\cite{11097694}. 
Their chain-of-thought reasoning enables them to combine traffic rules, spatial relationships, behavioral cues, and commonsense knowledge into a coherent decision-making process \cite{cai2024driving, wang2024omnidrive}. Crucially, LLMs possess strong zero-shot and few-shot generalization abilities \cite{suris2023vipergpt, yang2024doraemongpt}, making them promising for handling previously unseen interaction patterns in structured tactical driving scenarios.


Despite this progress, existing LLM-based driving frameworks largely focus on structured vehicle–vehicle interactions or rely on domain-specific training pipelines, leaving pedestrian-centric safety underexplored. Works on pedestrian prediction \cite{elallid2022dqn} highlight the difficulty of modeling human behavior, yet remain constrained by the limits of their training data or perceptual modalities. No prior work provides a comprehensive LLM-driven mechanism explicitly designed to anticipate abnormal, rare, or unseen pedestrian behaviors and translate them into safe tactical driving actions.

This paper builds on our preliminary study on LLM-augmented driving behavior planning~\cite{10.1145/3737611.3776953} and substantially expands it with a full pedestrian-centered framework, episodic memory retrieval, abnormal and unseen behavior evaluations, and more extensive comparisons with RL baselines.
To address these challenges, we propose an LLM-enhanced decision-making framework for AVs operating in pedestrian-rich environments. The framework translates structured environmental observations, such as positions, velocities, and lane data, into natural-language prompts processed by an LLM. The model then generates high-level tactical decisions, which are translated into control commands through a motion controller that enforces kinematic constraints. This design enables interpretable, human-aligned behavior without requiring handcrafted rules or dense supervision.

The proposed system emphasizes cautious and socially compliant interactions with pedestrians. By leveraging the LLM’s general world knowledge and contextual reasoning, the vehicle can anticipate potential pedestrian movements and react adaptively, reducing unsafe or abrupt maneuvers. Pedestrian interactions are among the most safety-critical challenges in AV deployment, as pedestrians are legally prioritized and their motion is inherently unpredictable. Failures in these scenarios not only increase accident risk but also erode public trust in autonomous systems.

Experimental evaluations in simulated single- and multi-pedestrian crossing scenarios demonstrate that the LLM can make safety-oriented tactical decisions even without task-specific fine-tuning. The model generalizes to unseen pedestrian interaction patterns and produces interpretable decisions, often outperforming RL baselines in adaptability and contextual awareness. Few-shot memory on limited historical driving experiences further improves stability and consistency. These results highlight the potential of language-based reasoning as a foundation for transparent, generalizable, and human-aligned autonomous driving behavior.

\color{black}


\section{Related Work}

\subsection{Reinforcement Learning-based Method}
Traditionally, AV control systems have been developed using reinforcement learning (RL) frameworks\cite{papini2021reinforcement, kuutti2020survey}. In general, reinforcement learning is a form of trial-and-error learning in which agents, such as AVs, learn optimal driving behaviors by interacting with their environment. Rather than relying on labeled data, the system is guided by a reward mechanism: the agent receives feedback based on the consequences of its actions (e.g., avoiding collisions), allowing it to improve its policy over time through the maximization of cumulative rewards.


Chen et al.\cite{9216550} propose a conditional Deep Q-Network that integrates global-path conditioning, a fuzzy-logic module, and a Conv-LSTM-based encoder to capture environmental information and stabilize end-to-end steering and acceleration control. 
Zhu et al.\cite{zhu2020safe} introduce a deep deterministic policy gradient-based car-following controller that learns directly from real trajectories and a simulation loop, optimizing a tri-objective reward encompassing safety, efficiency, and comfort by designing the actor and critic deep learning networks.

Lu et al. \cite{lu2020hierarchical} proposed a two-level control scheme for AV motion planning: a “decision” layer selects actions such as speeding up or changing lanes, and an “execution” layer converts them into safe, feasible paths. This structure ensures strategic plans remain within the vehicle’s capabilities, bridging high-level decisions with real-world motion.
Pathare et al.\cite{pathare2024tactical} introduce a tactical-level driving agent that not only chooses high-level manoeuvres (keep lane, change lane, accelerate, brake) via deep reinforcement learning but also quantifies how confident it is in each choice. By training an ensemble of Q-networks augmented with randomized prior functions (RPF), the agent can flag unfamiliar situations and fall back to a conservative action when its uncertainty is high, outperforming a standard Double-DQN baseline on simulated highways and avoiding crashes in out-of-distribution scenarios. 

\textit{Traditional RL pipelines for autonomous driving have been analyzed in depth, yet they remain sample-inefficient and slow to train, often requiring millions of simulation steps to converge to a stable policy. 
Because these agents overfit to the environment on which they were trained, their performance degrades sharply when the scene distribution shifts.
Seeking faster generalization, recent studies have replaced RL with trajectory-generation models based on diffusion processes\cite{liao2025diffusiondrive}, yet these models still rely on large domain-specific datasets and substantial offline training time.}

\subsection{LLM-based AV control}
The advent of LLMs has significantly expanded and enhanced various applications, particularly in the realm of AV systems. Researchers are increasingly focusing on the contextual understanding and reasoning capabilities of LLMs, especially their proficiency in interpreting and comprehending input text. When applied to the dynamic environments surrounding AV, LLMs demonstrate potential in facilitating planning and decision-making processes.
Beyond autonomous driving, LLM-based agent simulation has also been explored in our prior work for generating daily activities of smart-home simulator agents, suggesting the broader potential of language models for modeling human-centered behavior in simulated environments~\cite{10599909}.

Recent studies have explored integrating LLMs into AV frameworks to improve decision-making and planning. 
DriveMLM\cite{wang2023drivemlm} tokenizes temporal camera images, LiDAR sweeps, system-level traffic rules, and user commands, fuses them via a Vision-Transformers, and employs a LLaMA-based decoder to predict discrete path-and-speed planner states together with natural-language justifications, enabling closed-loop control in the CARLA simulator.
LLaDA\cite{li2024driving} focuses on adapting driving policies to regional traffic rules expressed in natural language, enabling rule-compliant decision-making across domains.
RDA-Driver\cite{huang2024making} has revealed a reasoning–decision misalignment in vision-language planners, showing that persuasive Chain-of-Thought (CoT) explanations can accompany unsafe trajectories, and counter this flaw with RDA-Driver, a bird’s-eye-view multimodal LLaMA that unifies reasoning and motion by aligning reasoning and decision by LLM.
Agent-Driver\cite{mao2023agent} treats an LLM as a cognitive scheduler. Several tools, such as getting occupancy on roads or predicting other vehicles' trajectories,  are installed in their system, and LLM chooses and uses tools, retrieves commonsense memories, runs CoT reasoning, and outputs a trajectory, while making every decision textually interpretable.
AccidentGPT\cite{wang2023accidentgpt} is the first large-scale system to jointly exploit multi-vehicle V2X, bird’s-eye-view perception, and GPT-level reasoning, enabling a unified model to forecast risks, issue warnings, and analyze accident causes for both autonomous and human-driven vehicles.
\textcolor{black}{ConvoyLLM\cite{lu2025convoyllm} equips each connected vehicle with a few-shot LLaMA agent, augments it via a task-oriented shared-memory pool, and couples the resulting high-level commands with an interlaced dynamic-graph planner, enabling robust obstacle avoidance, merging, splitting, and escort re-formation across multi-lane highways.}

\textit{These studies typically presuppose reliable relative-position information for neighboring vehicles, may require substantial multi-modal training data that limits cross-domain generalization, and pay scant attention to pedestrian interactions, necessitating retraining or data augmentation when such scenarios are critical. In contrast, our proposed method leverages the zero-shot capabilities of LLMs to handle pedestrian interactions and novel environments without additional training.}

\subsection{Pedestrian Safety Mechanism}
Ensuring pedestrian safety is essential for the real-world deployment of AVs. Unlike AVs, pedestrians are not integrated into computational systems, necessitating that AVs possess robust perception and understanding capabilities to interact safely with them. Consequently, designing trustworthy safety mechanisms is essential.

Trumpp et al.\cite{trumpp2022modeling} address the challenge of enabling AVs and pedestrians to safely interact at unsignalised crosswalks. They model it as a deep multi-agent reinforcement learning game, allowing both agents to learn and adapt. Their PCAM(pedestrian crash avoidance mitigation) policy nearly eliminates collisions and remains robust under noise and unpredictable behavior.
Elallid et al.\cite{elallid2022dqn} propose a Deep Q-Network (DQN) policy for autonomous driving in the CARLA simulator. The agent uses front-camera input and selects from discrete throttle, brake, and steering actions. Training rewards task completion, penalizes collisions, and encourages efficient, safe driving.
GPT-4V\cite{huang2024gpt} evaluates whether a GPT-4 Vision model can reason about, and forecast, pedestrian crossing behavior from raw images. Tested zero-shot on two public datasets, GPT-4V attains 57\%–67\% accuracy without any fine-tuning, exhibits strong commonsense explanations, yet still trails task-specific CNN/LSTM baselines and struggles with small or fast-moving pedestrians. 

\textit{In the context of autonomous driving using LLMs, methods explicitly prioritizing pedestrian safety in driving control remain unexplored. Even in studies employing bird’s-eye view perspectives, most rely on simplistic social force models or time-consuming training schemes that impose implicit penalties on the control model. In contrast, our work distinguishes itself by leveraging language models to explicitly prioritize pedestrian safety, setting it apart from existing research.}

\begin{figure*}[!tbp]
    \centering
    \includegraphics[width=0.98\linewidth]{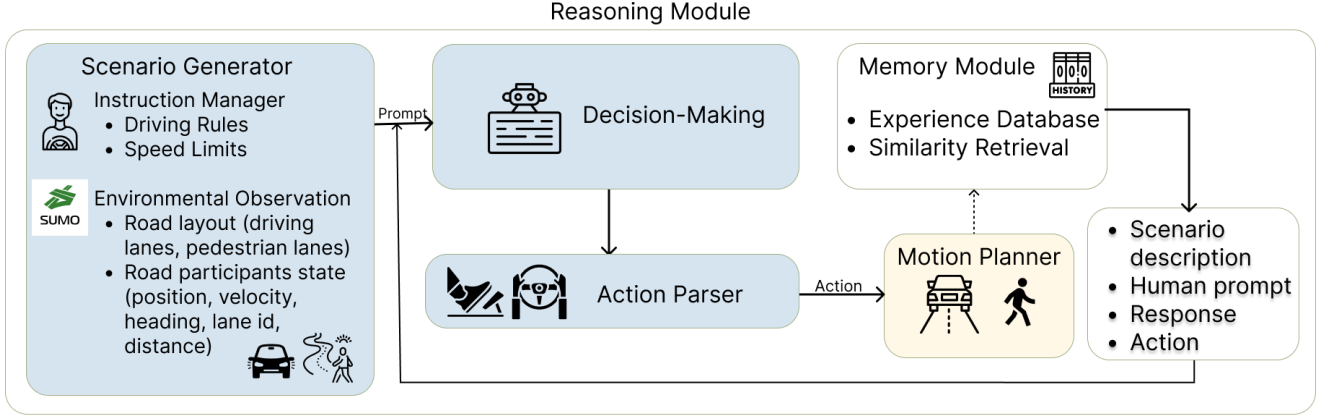}
    \caption{System overview.}
    \label{fig:system_overview}
\end{figure*}

\section{System Architecture}

\begin{figure*}[!tbp]
    \centering
    \includegraphics[width=0.98\linewidth]{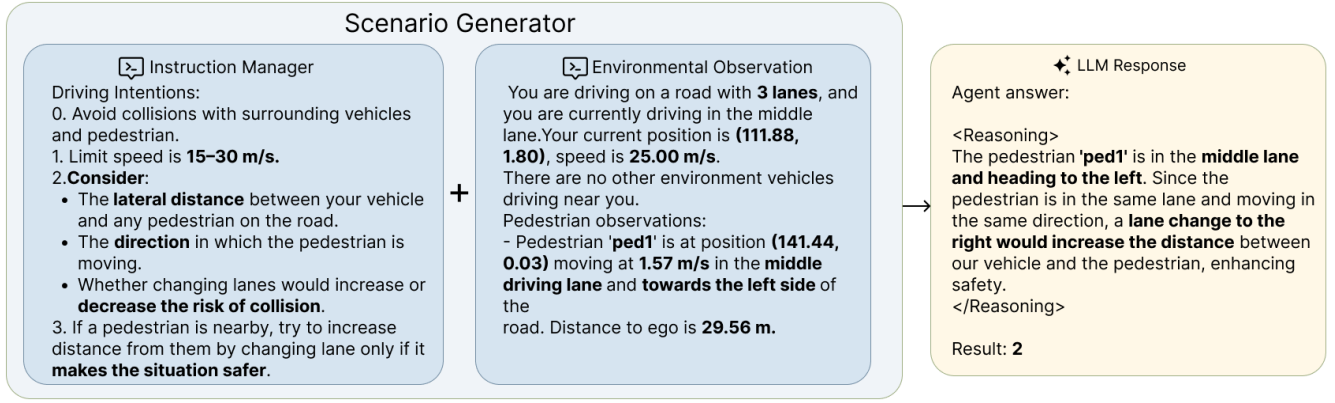}
    \caption{Example of scenario prompt and LLM response.}
    \label{imgs:prompt}
\end{figure*}

The proposed framework integrates an LLM-based Reasoning Module as the core of the AV decision pipeline, enabling interpretable, context-aware planning in complex urban environments. As illustrated in Figure~\ref{fig:system_overview}, the system consists of four main components: the Scenario Generator, Decision-Making, Action Parser and Motion Planner, and the Memory Module.
\paragraph{Closed-Loop and Horizon Execution}
The proposed framework operates as a receding-horizon closed-loop decision system. 
At each timestep $t$, \textit{the Scenario Generator} converts the current environment state into a structured textual description, which is provided to the LLM-based \textit{Decision-Making Module}. The LLM produces a high-level tactical action that is parsed and executed by \textit{the Motion Planner} over an execution horizon of $\Delta t$ simulation steps.
To reduce decision latency and maintain real-time responsiveness, LLM inference for the next control cycle is parallelized while the current action is being executed. Specifically, multiple LLM calls are dispatched concurrently using thread-based execution, enabling simultaneous reasoning over the upcoming state horizon. 
Once the simulator advances and the new state becomes available, the pre-computed decisions are collected and applied in the subsequent control cycle.
This pipelined perception--reasoning--action loop repeats throughout the episode, allowing the system to approximate receding-horizon planning while maintaining continuous environmental feedback.
\paragraph{The Scenario Generator} This component constructs the textual prompt that drives the LLM’s reasoning process and integrates two internal modules: the Environment Observation module and the Instruction Manager. The Observation module functions as the perception system, collecting structured data from the simulation, including road geometry, ego-vehicle state, and nearby agents such as vehicles and pedestrians. At each discrete time step~$t$, the ego-vehicle state is defined by its position $(x_t, y_t)$, speed $v_t$, heading $\theta_t$, and lane index $\text{lane}_t$, and is represented as
$s_t^{\text{ego}} = (x_t, y_t, v_t, \theta_t, \text{lane}_t)$.
This information is combined with dynamic attributes of surrounding agents (e.g., their velocity and heading) and traffic rules (e.g., speed limits), forming a structured and context-rich description of the driving scene.
The Instruction Manager defines the current driving goal or intent, such as maintaining lane, slowing down, or yielding to pedestrians. 
Prompts may optionally include a few-shot examples retrieved from \textit{the Memory Module} to enhance generalization and stability in unfamiliar conditions, whose process is shown in Figure~\ref{imgs:prompt}.

\paragraph{The Decision-Making component} This component processes each prompt through the LLM, which reasons over the provided context to produce a high-level tactical decision expressed in natural language, such as ``slow down due to pedestrian ahead'' or ``change to the left lane.'' This approach enables interpretable decision-making that can be analyzed and verified post hoc.
\paragraph{The Action Parser} This component implements a deterministic mapping $\mathcal{M}: T \rightarrow \mathcal{A}$,where $T$ denotes the LLM-generated text output and $\mathcal{A}=\{0,1,2,3,4\}$ represents the discrete action space. The action space $\mathcal{A}$ corresponds to: 0: Stay Idle, 1: Turn Left, 2: Turn Right, 3: Accelerate, 4: Decelerate.
\textit{The Action Parser} and \textit{Motion Planner} translate the LLM’s textual output into discrete control actions, including maintaining speed, accelerating, decelerating, turning left, and turning right. These actions are mapped to physically feasible motion primitives that respect the vehicle’s kinematic constraints and safety limits. The ego vehicle operates within a discrete action space consisting of five maneuvers: stay idle, turn left, turn right, accelerate, and decelerate.
The low-level motion controller executes these commands in real time, ensuring smooth transitions and stable control.


\paragraph{Memory Module.}
The Memory Module supports decision-making by storing structured numerical representations of previous successful examples between the ego vehicle and pedestrians. 
Instead of retaining full textual demonstrations, each memory entry is encoded as a normalized feature vector
\[\mathbf{x} = [dx,\, dy,\, v_{\text{ego}},\, v_{\text{ped}},\, \mathrm{TTC}],\]
representing relative longitudinal distance, lateral offset, ego velocity, pedestrian velocity, and time-to-collision.
The feature weights are empirically defined to emphasize safety-critical factors. 
In particular, the influence of lateral offset is increased in near-conflict situations, allowing the retrieval process to prioritize interactions with higher collision risk.

A kNN search retrieves the most similar past situations in the normalized feature space. 
The retrieved samples are subsequently de-serialized into natural-language templates containing scenario descriptions, reasoning traces, and selected actions, and are provided to the LLM prompt as few-shot demonstrations. 
Using compact numerical vectors instead of raw textual retrieval reduces inference latency while ensuring retrieval based on geometric and dynamic similarity rather than lexical matching.

\color{black}

\paragraph{The Motion Planner} applies a simplified nonholonomic kinematic model consistent with De Luca et al.~\cite{de2006kinematic}.
The vehicle is treated as a rigid body constrained to move along its heading direction, preventing instantaneous lateral displacement. The vehicle state evolves over a time interval~$\Delta t$ according to
\begin{equation}
\label{eq:kinematics_compact}
\begin{pmatrix}
x_{t+1} \\[2pt]
y_{t+1} \\[2pt]
\theta_{t+1} \\[2pt]
v_{t+1}
\end{pmatrix}
=
\begin{pmatrix}
x_t + v_t\cos\theta_t\,\Delta t \\[2pt]
y_t + v_t\sin\theta_t\,\Delta t \\[2pt]
\theta_t + k_\theta(\theta_{\text{target}}-\theta_t) \\[2pt]
v_t + a\,\Delta t
\end{pmatrix},
\end{equation}
where the acceleration command is
\begin{equation}
a = \min\left(a_{\max}, \max\left(-a_{\max}, \frac{v_{\text{target}} - v_t}{\Delta t}\right)\right) 
\end{equation}
with $v_{\text{target}}$ and $\theta_{\text{target}}$ derived from the LLM’s commonsense reasoning, and $a_{\max}$ denoting the physical acceleration bound. This ensures that the acceleration remains within the physically feasible range $[-a_{\max}, a_{\max}]$.


\color{black}

\color{black}

\section{Experimental Setup and Results}

\begin{figure}[!tbp]
    \centering
    \includegraphics[width=0.98\linewidth]{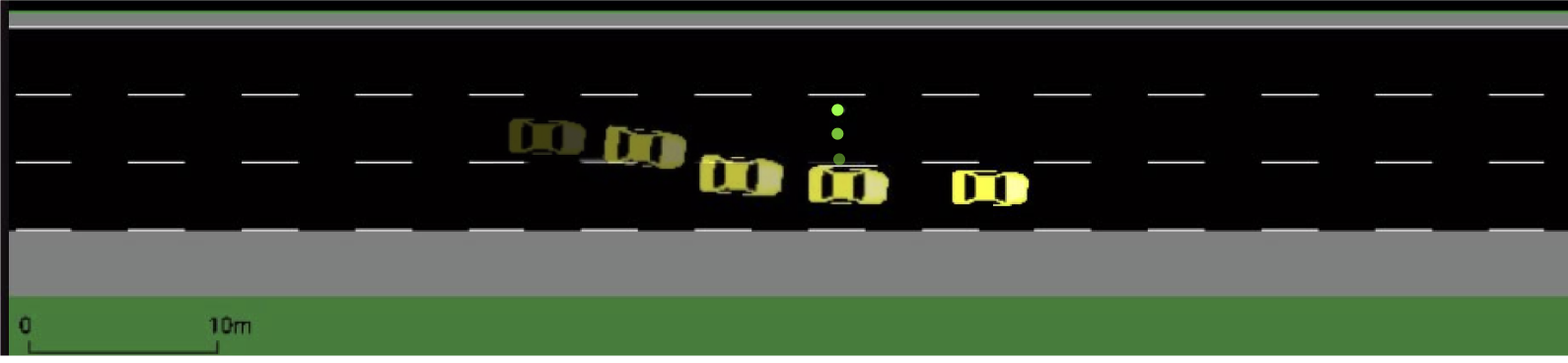}
    \caption{Historical trajectory of ego vehicle and pedestrian. The pedestrian is represented as a green dot.}
    \label{fig:LLM_trajectory}
\end{figure}

\subsection{Simulator and Environment}

All experiments were conducted using the SUMO microscopic traffic simulator (version 1.12.10). The environment models a three-lane urban roadway with bidirectional pedestrian access, enabling controlled evaluation of tactical driving behaviors including lane keeping, speed adaptation, and pedestrian avoidance. All positions and distances are represented in the SUMO global Cartesian coordinate system (meters).

Simulations were executed using fixed configuration files defining road geometry, vehicle dynamics, and pedestrian behavior policies. To ensure statistical robustness, each scenario was evaluated across 50 randomized seeds, where each seed generated distinct pedestrian spawn positions and motion trajectories.
Experiments were performed on a workstation equipped with an NVIDIA GeForce RTX 2080 Ti GPU (11 GB VRAM) and a multi-core CPU with 64 GB RAM.

\subsection{Evaluation Metrics}

Performance was evaluated using the following metrics:

\begin{itemize}
\item Success Rate: percentage of episodes completed without collision.
\item Average Speed: mean ego-vehicle speed during successful episodes.
\item Steps: number of simulation steps required to complete the scenario.
\item Minimum Pedestrian Distance: closest Euclidean distance between vehicle and pedestrian.
\item Minimum Lateral Pedestrian Distance: minimum lateral clearance between the vehicle body and pedestrians.
\item Minimum Time-to-Collision (TTC): smallest predicted collision time observed during an episode.
\end{itemize}


\subsection{Scenario and Memory Definitions}
\textbf{Seen Scenario:}
A pedestrian behavior type is included in the memory construction phase and therefore represented in the memory bank.

\textbf{Unseen Scenario:}
A pedestrian behavior type not included in the memory construction phase.

\textbf{Zero-Shot Setting (No Memory):}
The LLM-based agent operates without episodic memory retrieval. Decisions are generated solely from the current scene description and driving rules, without access to prior interaction examples.

\textbf{Few-Shot Setting (With Memory):}
The LLM-based agent is equipped with an episodic memory bank constructed from prior interactions. During inference, relevant memory entries are retrieved and incorporated into the prompt to guide decision-making.

The Turn-back scenario, described later in Sections~\ref{sec:memory_in_seen_scenario} and ~\ref{sec:multi_pedestrian_scaling}, uses a memory bank containing 114 state vectors, while the single pedestrian crossing memory bank contains 82. 
Each vector represents one decision-step state stored for $k$-NN retrieval, rather than an entire episode trajectory.

\subsection{Models}

We evaluate four instruction-following large language models: Gemma3-1B, Qwen3-1.7B, LLaMA3.2-1B, and Gemma3-27B. All LLMs operate without gradient updates during evaluation.
Inference for all LLMs was performed using a temperature of 0.0, top-p of 0.9, and a maximum generation length of 2000 tokens to ensure deterministic and comparable decision outputs.
As a RL baseline, we include a Deep Q-Network (DQN) agent trained for 200,000 timesteps in the same environment. The DQN policy is evaluated without further training.
All experiments were conducted over 50 independent episodes per experimental condition.
Table~\ref{tab:success_rates} summarizes the collision-free success rates obtained across all research questions and experimental settings, including single-pedestrian unseen scenarios, memory-augmented evaluations in seen environments, increasing scene complexity with multiple pedestrians, and cross-behavior memory transfer tests.

\begin{table*}[t]
\centering
\caption{Collision-Free Success Rate Across Experiments}
\label{tab:success_rates}
\begin{tabular}{lllc}
\toprule
\textbf{Exp} & \textbf{Scenario} & \textbf{Model} & \textbf{Success Rate (\%)} \\
\midrule

\multirow{3}{*}{Exp1}
& \multirow{3}{*}{Single Pedestrian (Unseen)}
& \textbf{Gemma3-27B (No Memory)} & \textbf{68.0} \\
& & Dueling DQN & 17.7 \\
& & Randomized Prior Functions (RPF) & 42.8 \\
\midrule

\multirow{7}{*}{Exp2}
& \multirow{7}{*}{Single Pedestrian (Seen)}
& Gemma3-1B (No Memory) & 72.0 \\
& & Qwen3-1.7B (No Memory) & 74.0 \\
& & LLaMA3.2-1B (No Memory) & 76.0 \\
& & \textbf{Gemma3-1B (With Memory)} & \textbf{96.0} \\
& & Qwen3-1.7B (With Memory) & 92.0 \\
& & LLaMA3.2-1B (With Memory) & 68.0 \\
& & DQN & 82.0 \\
\midrule

\multirow{4}{*}{Exp3}
& \multirow{4}{*}{Three Pedestrians}
& Gemma3-1B (With Memory) & 88.0 \\
& & Qwen3-1.7B (With Memory) & 76.0 \\
& & LLaMA3.2-1B (With Memory) & 68.0 \\
& & \textbf{DQN} & \textbf{92.0} \\
\midrule

\multirow{3}{*}{Exp4}
& Turn-back (Seen)
& \textbf{Gemma3-1B (With Memory)} & \textbf{94.0} \\
\cmidrule{2-4}
& Hesitation (Unseen)
& \textbf{Gemma3-1B (With Memory)} & \textbf{82.0} \\
\cmidrule{2-4}
& Bidirectional Crossing (Unseen)
& \textbf{Gemma3-1B (With Memory)} & \textbf{90.0} \\
\bottomrule
\end{tabular}
\end{table*}

\subsection{Zero-Shot LLM vs. RL Baselines without Episodic Memory}

This experiment evaluates whether LLM agents can generalize to previously unseen pedestrian interactions without relying on external memory. The objective is to establish a fair zero-shot comparison between reasoning-based decision making and reinforcement learning (RL) policies trained under limited environmental assumptions.

We compare a zero-shot LLM agent (Gemma-27B) with two tactical driving baselines, Dueling DQN and Randomized Prior Functions (RPF)~\cite{hoel2020tacticaldecisionmakingautonomousdriving,pathare2024tactical}. RL agents were trained exclusively on vehicle–vehicle interaction tasks and were never exposed to pedestrian behaviors during training. Similarly, the LLM operates without memory augmentation or task-specific adaptation.
Exp1 does not evaluate which approach is superior in general. Rather, it tests a specific claim in the literature — that tactically-trained RL agents can generalize to unseen interaction types. We evaluate this claim by placing these agents in a pedestrian scenario outside their training distribution, alongside a zero-shot LLM that also has no task-specific adaptation. This is an out-of-distribution generalization test, not a head-to-head competition on equal footing.

The zero-shot LLM achieves a collision-free success rate of 68\%, outperforming RPF (42.8\%) and DQN (17.7\%). RL agents frequently collided or halted prematurely, indicating limited transfer beyond their training distribution. In contrast, the LLM adapts speed and trajectory based on contextual reasoning despite the absence of prior pedestrian experience.

In terms of safety behavior, the LLM maintained a median minimum lateral clearance of approximately 3.3~m while completing most episodes. RL agents preserved substantially larger clearances (approximately 5~m) but often failed to complete the scenario due to excessive braking or conservative stopping policies. This contrast highlights a qualitative difference in decision strategies: the LLM executes tighter yet confident avoidance maneuvers informed by contextual reasoning, whereas RL agents apply rigid avoidance behaviors that prioritize separation at the cost of task completion.

These results indicate that reasoning-based policies exhibit stronger zero-shot generalization to unseen agent types compared to trained RL baselines.











\subsection{Episodic Memory Effects in Seen Scenarios}\label{sec:memory_in_seen_scenario}

This experiment investigates whether episodic memory improves robustness in previously observed pedestrian-crossing scenarios, evaluated using success rate, minimum ego–pedestrian distance, and minimum lateral clearance. 

\begin{figure}[t]
    \centering
    \includegraphics[width=0.8\linewidth]{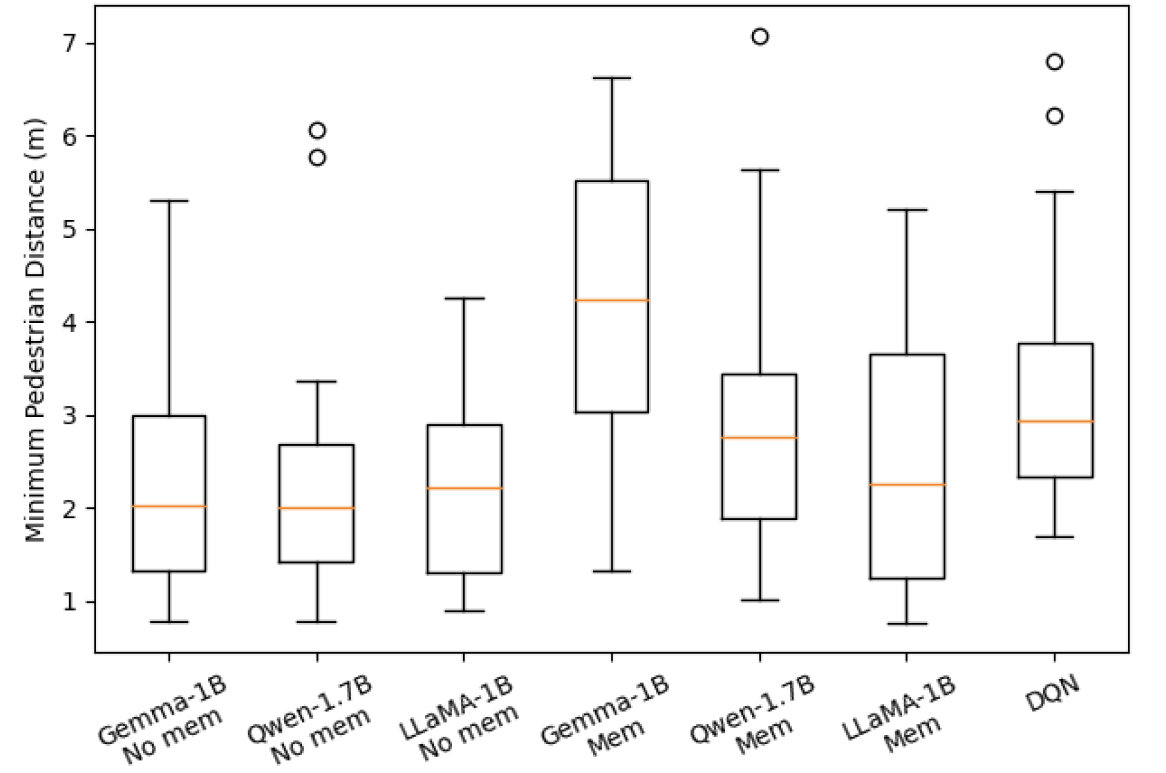}
    \caption{Exp2: Min Pedestrian Distance Distribution (Single Pedestrian, Seen).}
    \label{imgs:RQ2_Min_Dist}
\end{figure}

\begin{figure}[t]
    \centering
    \includegraphics[width=0.8\linewidth]{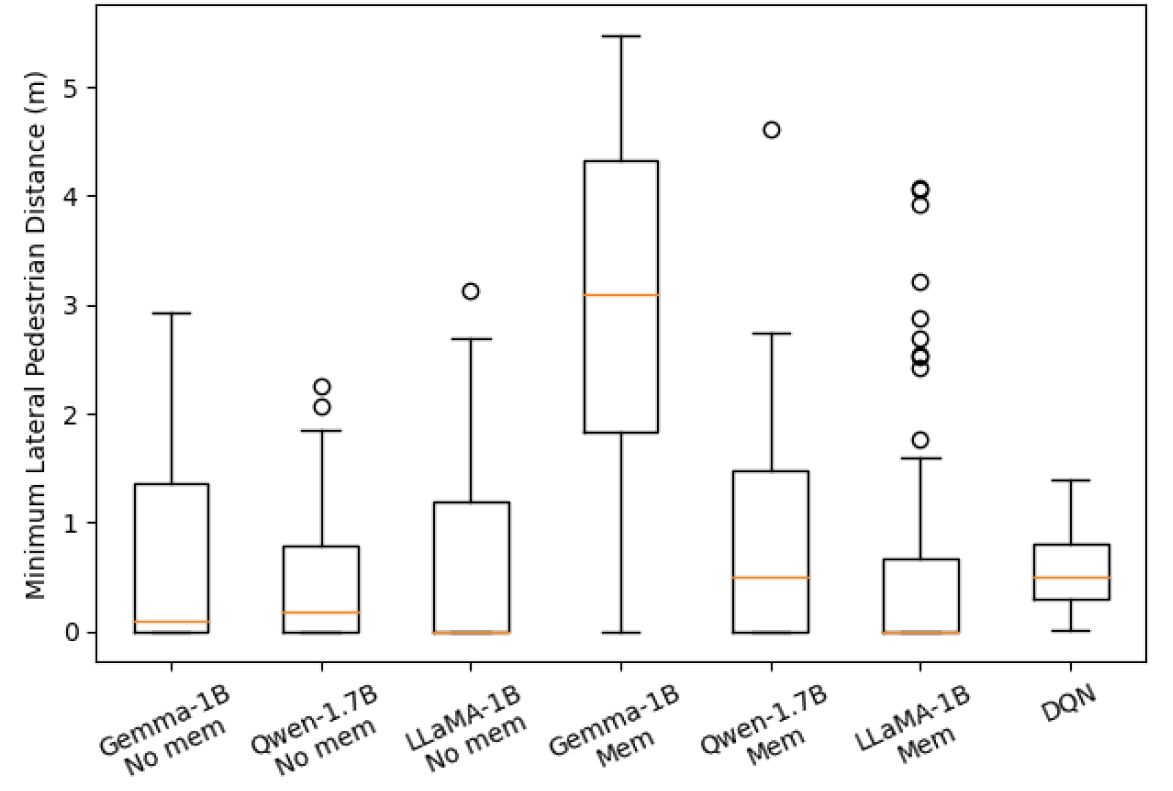}
    \caption{Exp2: Min Lateral Pedestrian Distance Distribution (Single Pedestrian, Seen).}
    \label{imgs:RQ2_Min_Lat_Dist}
\end{figure}

For Gemma-1B, memory increases the success rate from 72.0\% to 96.0\%, minimum ego–pedestrian distance from 2.63 m to 4.34 m, and minimum lateral clearance from 0.96 m to 3.17 m (+2.21 m). These results indicate that memory augmentation consistently enlarges spatial buffers during pedestrian interaction.
The DQN baseline achieves an 82.0\% success rate but maintains a minimum lateral clearance of only 0.61 m, which is 2.56 m narrower than Gemma-1B with memory. This contrast highlights behavioral difference: the RL policy learns to complete the task within tight margins, while the memory-augmented LLM leverages contextual reasoning over retrieved experiences to maintain substantially larger spatial separation, reflecting an awareness of pedestrian proximity that extends well beyond minimum collision avoidance requirements. The distributions of the minimum ego–pedestrian distance and minimum lateral pedestrian distance are visualized in Fig.~\ref{imgs:RQ2_Min_Dist} and Fig.~\ref{imgs:RQ2_Min_Lat_Dist}, respectively.

Overall, episodic memory improves robustness in seen scenarios primarily by enlarging lateral safety margins, reflecting a shift toward defensive rather than minimally compliant collision avoidance. Whether these gains reflect genuine behavioral generalization or scenario-specific retrieval is examined in the following experiment.

\subsection{Scaling to Multi-Pedestrian Scenes}\label{sec:multi_pedestrian_scaling}

This experiment evaluates whether driving policies learned from single-pedestrian interactions can generalize to more complex multi-pedestrian environments. All memory-augmented LLM agents were constructed using episodic memories derived from the one-pedestrian walking scenario, while the DQN baseline was trained under the same single-pedestrian condition. Evaluation was performed in a multi-pedestrian setting, introducing increased spatial uncertainty and simultaneous collision risks.

The DQN agent achieves the highest task completion rate of 92.0\%. However, this performance is associated with minimal safety margins, yielding a mean minimum pedestrian distance of $2.593 \pm 1.057$~m and an extremely small minimum lateral distance of $0.034 \pm 0.029$~m, indicating collision avoidance behavior that prioritizes completion over spatial safety.

Memory-augmented LLM agents maintain competitive performance while preserving substantially larger safety buffers. Gemma-1B with memory achieves an 88.0\% success rate with a minimum pedestrian distance of $3.009 \pm 1.217$~m and a minimum lateral distance of $1.462 \pm 1.144$~m. Qwen-1B with memory attains a 76.0\% success rate while maintaining a minimum pedestrian distance of $2.758 \pm 1.256$~m and a lateral clearance of $1.276 \pm 1.180$~m. LLaMA-1B with memory achieves 68.0\% success with a minimum pedestrian distance of $2.364 \pm 1.348$~m and a lateral separation of $0.715 \pm 1.259$~m. Across all models, LLM-based policies consistently maintain lateral distances substantially larger than those produced by the DQN baseline.

Time-to-collision (TTC) measurements suggest behavioral differences in response timing. Memory-enabled agents maintain lower but stable TTC values (Gemma-1B: $0.351 \pm 0.554$~s, Qwen-1B: $0.231 \pm 0.481$~s, LLaMA-1B: $0.222 \pm 0.383$~s), reflecting earlier trajectory adjustment compared to the memory-free agent ($0.615 \pm 1.701$~s), which exhibits higher variance indicative of delayed or unstable reactions.

A notable exception is LLaMA3.2-1B, where memory augmentation lowers the success rate from 76.0\% to 68.0\%. We attribute this to the model's limited capacity to integrate retrieved context, where conflicting episodes introduce hesitation rather than guidance. This suggests memory augmentation is not uniformly beneficial — models below an instruction-following threshold may require retrieval filtering or summarization before episodic memory reliably improves performance.

Overall, the results indicate that while reinforcement learning achieves strong success rates under increased scene complexity, memory-augmented LLM agents generalize more conservatively, preserving larger interpersonal safety margins through contextual scene awareness rather than purely goal-driven optimization.

\subsection{Cross-Behavior Generalization}

This experiment evaluates whether experience acquired from one pedestrian behavior generalizes to distinct and previously unseen behavioral patterns in more complex scenarios. Memory entries were constructed exclusively using the turn-back pedestrian scenario, in which pedestrians reverse direction while crossing. Evaluation in this experiment is restricted to Gemma3-1B, as it demonstrated the most stable and consistent memory-augmented performance across Exp2 and Exp3, making it the most suitable candidate for isolating the effect of cross-behavior memory transfer independently of model-capacity variability. Evaluation was subsequently performed without updating memory on two unseen behaviors: hesitation and bidirectional crossing. 

These scenarios introduce substantially different interaction dynamics. Hesitation behavior produces unpredictable stop--go motion, increasing temporal uncertainty, and requiring continuous replanning. Bidirectional crossing introduces spatial ambiguity through competing pedestrian trajectories. Consequently, successful performance requires adaptive reasoning rather than direct recall of stored motion patterns.

Using memory derived solely from the turn-back scenario, the agent achieved a success rate of 94.0\% (95\% CI: 83.8\%--97.9\%) on the memory-matched turn-back evaluation across 50 runs. The agent maintained a mean minimum pedestrian distance of 3.76~m (95\% CI: 3.40--4.13~m) and a mean time-to-finish (TTF) of 3.21~s (95\% CI: 3.08--3.34~s).

Despite the absence of behavior-specific memories, performance remained high in unseen scenarios. The hesitation scenario achieved a success rate of 82.0\% (95\% CI: 69.2\%--90.2\%), reflecting the increased difficulty of safely resolving unpredictable pedestrian intent. The agent maintained a mean minimum distance of 3.27~m (95\% CI: 2.83--3.72~m) and a mean TTF of 3.19~s (95\% CI: 3.07--3.32~s).
The bidirectional scenario achieved a success rate of 90.0\% (95\% CI: 78.6\%--95.7\%), indicating stable adaptation under multi-directional interaction complexity, with a mean minimum distance of 2.94~m (95\% CI: 2.64--3.24~m) and a mean TTF of 3.15~s (95\% CI: 3.03--3.26~s). Safety margins and completion times remained broadly consistent across all behavioral conditions.

These results demonstrate that retrieved memories encode generalizable interaction principles rather than scenario-specific trajectories, enabling the agent to adapt previously learned avoidance strategies to unseen pedestrian behaviors without retraining or additional memory accumulation. This supports the broader claim that memory augmentation enables robust decision-making under diverse and unpredictable real-world interactions.


\begin{figure}[t]
    \centering
    \includegraphics[width=0.8\linewidth]{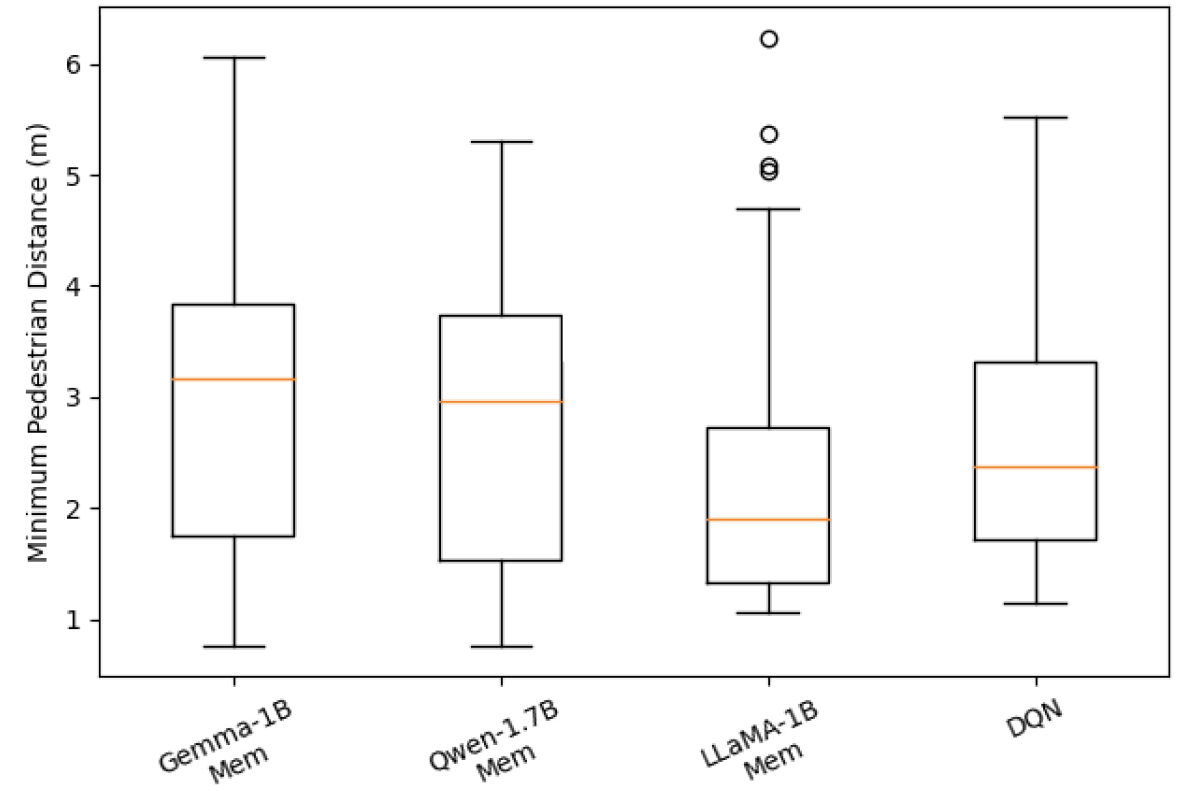}
    \caption{Exp3: Min Pedestrian Distance Distribution.}
    \label{imgs:RQ3_Min_Dist.png}
\end{figure}
\begin{figure}[t]
    \centering
    \includegraphics[width=0.8\linewidth]{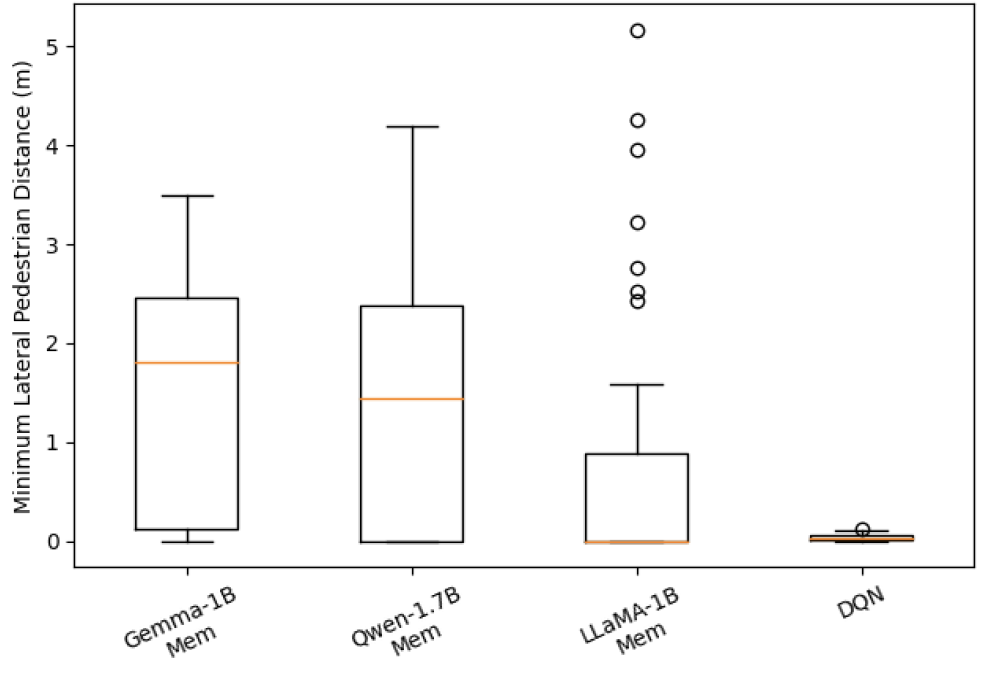}
    \caption{Exp3: Min Lateral Pedestrian Distance Distribution.}
    \label{imgs:RQ3_Min_Lat_Dist}
\end{figure}


\section{Computation Efficiency}
Inference latency was evaluated to assess real-time applicability of the decision-making policies. The Dueling DQN agent demonstrates highly efficient execution, requiring approximately 1.24–1.45 ms per decision step with minimal variance, reflecting the low computational cost of neural network forward inference. In comparison, the Gemma3-1B LLM exhibits higher variability due to autoregressive reasoning, with decision-step inference times ranging from near-instant responses to a maximum of 1.26 s (standard deviation 0.19 s).

Despite the computational advantage of DQN, the inference latency of Gemma3-1B remains compatible with real-time tactical planning operating at low decision frequencies. This suggests that LLM-based agents can remain practically deployable while providing improved reasoning capability compared to conventional reinforcement learning policies.

\begin{table}[t]
\centering
\caption{Inference time comparison between Gemma3-1B and Dueling DQN (per decision step).}
\begin{tabular}{lcccc}
\toprule
\textbf{Model} & \textbf{Avg (s)} & \textbf{Max (s)} & \textbf{Min (s)} & \textbf{Std (s)} \\
\midrule
Gemma3-:1B & 0.1178 & 1.2647 & 0.00006 & 0.1887 \\
DQN & 0.001306 & 0.001448 & 0.001240 & 0.000071 \\
\bottomrule
\end{tabular}
\label{tab:inference_times}
\end{table}

\section{Limitations}
Several limitations are acknowledged. First, the present evaluation focuses on tactical maneuver decision-making in the SUMO simulator. While this setting enables controlled and reproducible analysis of pedestrian-aware planning behavior, SUMO abstracts away several factors that are critical in full autonomous driving systems, including raw sensor perception, occlusions, perception noise, and limited field-of-view conditions. Therefore, validating the proposed framework in richer simulation platforms such as CARLA, and ultimately in real-world driving environments, remains an important direction for future work.
Second, the framework relies on structured state representations rather than end-to-end perceptual inputs. This design allows the study to isolate the reasoning and decision-making capability of LLM-based planning, but it limits direct applicability to complete autonomous driving pipelines that operate from camera, LiDAR, or multi-modal sensor data. Future work should integrate the proposed reasoning module with perception and prediction components, and evaluate its robustness under uncertain or imperfect state estimation.

\color{black}

\section{Conclusion}

This study introduced a framework that integrates LLMs into the decision-making process of AVs to enhance behavioral planning in pedestrian-rich environments. By translating structured scene observations into natural language prompts, the system enables interpretable, context-aware reasoning and generates tactical driving decisions aligned with human expectations. Experiments in the SUMO simulator, covering single- and multi-pedestrian jaywalking scenarios, demonstrated that LLM-based agents, especially those supported by few-shot memory, outperform RL baselines in collision avoidance, lateral clearance, and behavioral consistency. Unlike reactive RL policies trained on narrow objectives, LLM-based policies adapt to unseen pedestrian behavior patterns within the evaluated SUMO scenarios. The proposed framework also promotes socially interpretable driving behavior, prioritizing smoother lane changes, early evasive actions, and cautious maneuvers near pedestrians. Its natural language reasoning enhances transparency and facilitates post-hoc analysis of decision logic.
Overall, the results suggest that LLM-based reasoning can serve as a viable and interpretable component for tactical behavioral planning in structured pedestrian-interaction scenarios, offering a promising step toward more generalizable and human-aligned autonomous driving systems.

\bibliographystyle{IEEEtran}
\bibliography{_ref.MonoFi}

\end{document}